\documentclass[conference]{IEEEtran}
\IEEEoverridecommandlockouts

\usepackage{cite}
\usepackage{amsmath,amssymb,amsfonts}
\usepackage{algorithmic}
\usepackage{graphicx}
\usepackage{textcomp}

\usepackage{url}
\usepackage{makecell}
\usepackage[font=normalsize]{caption}

\usepackage{xcolor}
\def\BibTeX{{\rm B\kern-.05em{\sc i\kern-.025em b}\kern-.08em
    T\kern-.1667em\lower.7ex\hbox{E}\kern-.125emX}}
\begin{document}

\title{Gastroendoscopy View Synthesis:\\A New Real Dataset and Evaluation
{\footnotesize \textsuperscript{}}
\thanks{}
}

\author{Masaki Minai\IEEEauthorrefmark{1},
Yusuke Monno\IEEEauthorrefmark{1},
Sho Suzuki\IEEEauthorrefmark{2}, and
Masatoshi Okutomi\IEEEauthorrefmark{1}
\thanks{This study was supported by AMED under Grant Numbers JP24hma922022 and JP25hma322042. It was also supported by JSPS KAKENHI under Grant Numbers JP24K15772 and JP26K03332.}
\thanks{\IEEEauthorrefmark{1}M. Minai, Y. Monno, and M. Okutomi are with the Department of Systems and Control Engineering, School of Engineering, Institute of Science Tokyo, Meguro-ku, Tokyo 152-8550, Japan (email: \{mminai,ymonno,mxo\}@ok.sc.e.titech.ac.jp).}
\thanks{\IEEEauthorrefmark{2}S. Suzuki is with the Department of Gastroenterology, International University of Health and Welfare Ichikawa Hospital, Ichikawa-shi, Chiba 272-0827, Japan.}
}

\maketitle

\begin{abstract}
Novel view synthesis (NVS) is an active research topic in computer vision, owing to the success of neural radiance field (NeRF) and 3D Gaussian splatting (3DGS) methods. While NVS opens the door to potential applications in gastroendoscopy, such as extending the field of view of endoscopic images and enabling digital twins for 3D archiving and endoscopist manipulation training, the dataset is insufficient to evaluate NVS for gastroendoscopy. In this paper, we present the first real gastroscopy dataset for NVS, namely the GastroNVS dataset, which contains a set of gastroscopic images, camera poses, and a point cloud for real gastroendoscopy inspection. To assess the suitability of the GastroNVS dataset, we evaluate several 3DGS methods and discuss the challenges for future development. The dataset is available on request from our project page.
\end{abstract}

\begin{IEEEkeywords}
Gastroendoscopy, novel view synthesis, structure from motion, 3D Gaussian splatting, dataset.
\end{IEEEkeywords}

\section{Introduction}

Novel view synthesis (NVS) is the task of synthesizing arbitrary viewpoint images from real captured images of the scene. NVS is actively studied, especially after the development of neural radiance field (NeRF)~\cite{mildenhall2020nerf} and 3D Gaussian splatting (3DGS)~\cite{kerbl20233d} methods, demonstrating remarkable performance in the computer vision field. NVS also has potential applications in endoscopy, such as extending the field of view of endoscopic images and enabling digital twins for 3D archiving and endoscopist manipulation training.

Some studies have addressed NVS for colonoscopy~\cite{SHI2025131445,psychogyios2023realistic,bonilla2024gaussian,kaleta2025pr,wang2025moving} and gastroendoscopy~\cite{jiang2024neural,lu2025texturing} by applying NeRF-based~\cite{SHI2025131445,psychogyios2023realistic,jiang2024neural,lu2025texturing} or 3DGS-based~\cite{bonilla2024gaussian,kaleta2025pr,wang2025moving} methods. However, endoscopy datasets are still lacking for NVS. Table~\ref{tab1} summarizes publicly available endoscopy datasets applied to the evaluation of NVS in the past literature. The SCARED dataset~\cite{allan2021stereo} records porcine cavity and is originally designed for stereo depth estimation with ground-truth~(GT) depth maps. The EndoNeRF dataset~\cite{wang2022neural} records surgical scenes of human prostate. Each scene is captured from a fixed viewpoint over time and is aimed at addressing surgical tool occlusion removal rather than performing NVS from arbitrary viewpoints. The EndoMapper dataset~\cite{azagra2023endomapper} records colonoscopy data with the estimated camera poses, and the reconstructed point cloud by structure from motion~(SfM)~\cite{schonberger2016structure}. The C3VD dataset~\cite{bobrow2023colonoscopy} records phantom colon images, which are registered to the GT 3D model of the phantom, providing GT depth and normal maps. While this dataset is applied to evaluating NVS, the viewpoint change is relatively small for each scene. As far as we know, there is no existing public gastroendoscopy dataset for NVS with a large viewpoint change.

\begin{table*}[t]
\centering
\small
\caption{Comparison of endoscopic datasets for NVS.}
\label{tab1}
\renewcommand{\arraystretch}{1.15}
\setlength{\tabcolsep}{6pt}

\begin{tabular}{lcccccccc}
\hline
Dataset & Year & Organ &
\makecell{Viewpoint\\change} &
\makecell{Camera\\pose} &
\makecell{GT depth\\map} &
\makecell{GT normal\\map} &
\makecell{SfM point\\cloud} &
\makecell{Total\\scenes} \\
\hline

SCARED~\cite{allan2021stereo} 
& 2021
& Porcine stomach
& Large
& \checkmark 
& \checkmark 
& 
& 
& 9 \\

EndoNeRF~\cite{wang2022neural} 
& 2022
& Human prostate
& Fixed 
& \checkmark 
& 
& 
& 
& 6 \\

EndoMapper~\cite{azagra2023endomapper} 
& 2022
& Human colon 
& Large
& \checkmark 
& 
& 
& \checkmark 
& 96 \\

C3VD~\cite{bobrow2023colonoscopy}
& 2023
& Phantom colon 
& Small
& \checkmark 
& \checkmark 
& \checkmark
& 
&  20\\

GastroNVS (Ours)
& 2026
& Human stomach
& Large
& \checkmark 
& 
& 
& \checkmark 
& 5 \\
\hline
\end{tabular}
\end{table*}

\begin{table*}[t]
\centering
\small
\vspace{2mm}
\caption{Data acquisition settings and SfM reconstruction results for each sequence.}
\label{tab2}

\renewcommand{\arraystretch}{1.15}
\setlength{\tabcolsep}{4pt}

\begin{tabular}{lccccc}
\hline
 &  &  & \textbf{Sequence} &  &  \\
\hline
 & A & B & C & D & E \\
\hline
\multicolumn{6}{l}{\textbf{Data acquisition}} \\
\hline
Date   & 06/08/2024 & 16/08/2024 & 06/08/2025 & 27/05/2025 & 05/08/2025 \\
System & ELUXEO7000 & ELUXEO7000 & ELUXEO8000 & LASEREO7000 & ELUXEO8000 \\
Scope  & EG-760Z    & EG-740N    & EG-860Z    & EG-L580RD7 & EG-760Z \\
\hline
\multicolumn{6}{l}{\textbf{Frame extraction}} \\
\hline
\# of frames & 518 & 433 & 366 & 257 & 432 \\
Lesion & \checkmark &            & \checkmark & \checkmark & \checkmark \\
\hline
\multicolumn{6}{l}{\textbf{SfM reconstruction}} \\
\hline
\# of reconstructed frames & 516 & 433 & 366 & 257 & 430 \\
\# of reconstructed 3D points & 82{,}145 & 68{,}176 & 67{,}109 & 51{,}031 & 45{,}035 \\
Mean reprojection error & 1.07384 & 1.36415 & 1.09455 & 1.07221 & 1.09844 \\
Image size after undistortion & 1120$\times$1128 & 1254$\times$1248 & 1101$\times$1069 & 1218$\times$1247 & 1153$\times$1153 \\
\hline
\end{tabular}

\renewcommand{\arraystretch}{1.0}
\end{table*}

In this paper, we introduce GastroNVS, the first real gastroendoscopy dataset for NVS. The dataset contains gastroendoscopic image sequences along with their estimated camera poses and the reconstructed 3D point cloud by SfM. The dataset was collected during a real gastroendoscopy inspection, which captures the gastric surfaces of real patients with gastric lesions and includes relatively large viewpoint changes that occurred in the real situation. Thus, it provides an evaluation platform for in-the-wild environments. Using the GastroNVS dataset, we evaluate several 3DGS methods to assess the suitability of the dataset for NVS and highlight the challenges revealed by the experiments. The dataset is available on request from our project page\footnote{Project page: \textcolor{blue}{\url{http://www.ok.sc.e.titech.ac.jp/res/GastroNVS/}}}.

\section{Dataset Construction}

\subsection{Ethics}
This study was conducted in accordance with the Declaration of Helsinki. The Institutional Review Board at International University of Health and Welfare approved the study protocol and the data collection (Approval No. 25-CI-005). Informed consent was obtained from patients before collecting endoscopic images. This study was also approved by the Research Ethics Committee of Institute of Science Tokyo (Approval No. 2025197) to process the collected data.

\subsection{Data Acquisition}
We used endoscopic video recordings from five subjects. By following the observation in~\cite{widya2019whole}, the indigo carmine blue dye, which is widely used in chromoendoscopic examinations, was sprayed onto the gastric surfaces in advance of the recordings to enhance the visibility of the gastric structures and facilitate robust feature extraction and matching in SfM. The endoscopic systems and scope types used are summarized in the top part of Table~\ref{tab2}, where the examinations were conducted using Fujifilm monocular endoscopic systems, specifically ELUXEO 7000, ELUXEO 8000, or LASEREO 7000, along with Fujifilm scopes as listed in the table. All video recordings were obtained at the frame rate of 30~fps.

\subsection{Frame Extraction}
\begin{figure}[t]
\centering
\includegraphics[width=0.9\columnwidth]{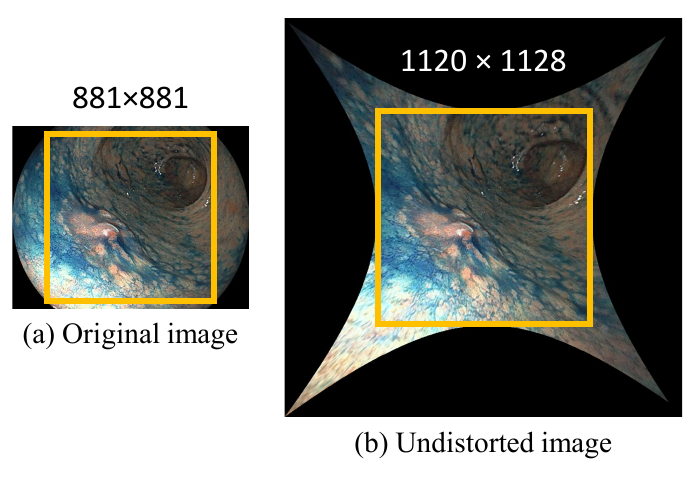}\\ \vspace{-2mm}
\caption{(a) The cropping of the original image to input to SfM and (b) the cropping of the undistorted image after the undistortion.}
\label{fig1}
\end{figure}

From the captured videos, we extracted subsequences of approximately 10–20 seconds per subject that capture wide regions of the stomach. Scenes without excessive surface deformation were selected, although some sequences still include subtle surface deformations. Then, we decomposed the video into individual image frames and removed frames that could potentially degrade the quality of subsequent reconstructions. For example, the frames in which the blue dye was spread to the wide regions of the image and the frames in which the camera was positioned extremely close to the gastric surface were excluded. Then, we cropped the original video frame as shown in Fig.~\ref{fig1}(a) to remove the black regions, resulting in a uniform resolution of 881 × 881 for all sequences. 

\begin{figure*}[t]
\centering
\includegraphics[width=\textwidth]{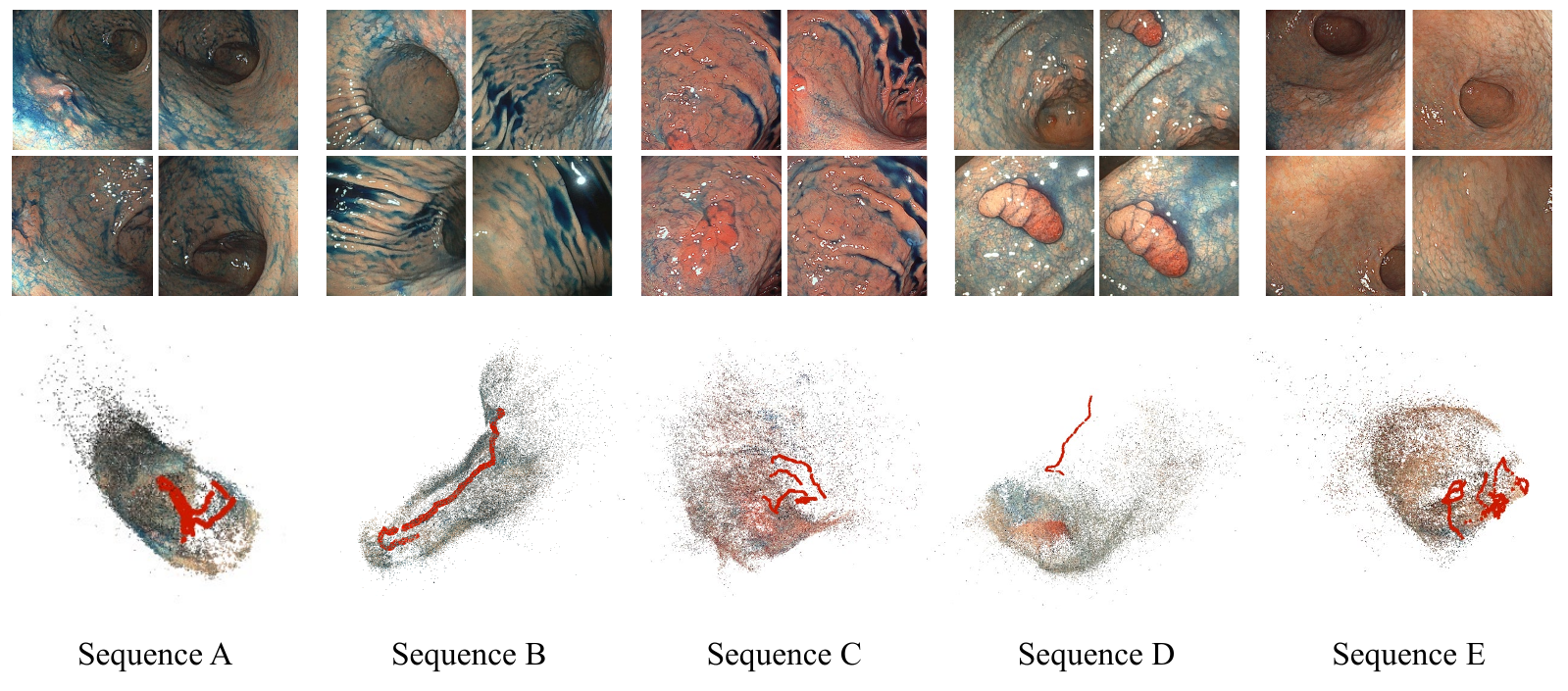}
\caption{Examples of the input images to SfM and the reconstructed 3D point cloud for each sequence.}
\label{fig2}
\end{figure*}

As shown in the middle part of Table~\ref{tab2}, the number of extracted frames ranged from 257 to 518, with four sequences containing gastric lesions. No occlusion by the endoscope was present in the field of view in all the sequences. Some representative frames from each sequence are shown in the top part of Fig.~\ref{fig2}.

\subsection{SfM}
Using each image sequence described in the previous section, we performed SfM using a common software, COLMAP~\cite{schonberger2016structure}. Specifically, the extraction of the SIFT features~\cite{lowe2004distinctive} from each image is firstly performed, followed by matching these features across images by exhaustive matching. Then, the camera poses and the 3D point cloud were estimated through triangulation and global bundle adjustment. Simultaneously, the intrinsic camera parameters of the endoscope were estimated using the radial fisheye camera model, which is suitable for the endoscope with a wide viewing angle. This model represents lens distortion using second- and fourth-order radial terms. 

Since NeRF and 3DGS methods are typically designed under the assumption of pinhole camera representations, we converted the original fisheye camera representation into an equivalent pinhole camera representation by compensating for lens distortion based on the intrinsic camera parameters estimated in SfM. This undistortion process outputs distortion-free images, whose sizes are decided so as to exclude the black regions, as shown in Fig.~\ref{fig1}(b).

\subsection{Dataset Summary}
The resulting point cloud of each sequence is shown in the bottom part of Fig.~\ref{fig2}. The quantitative results of SfM are reported in the bottom part of Table~\ref{tab2}. For three of the sequences, the camera poses were successfully reconstructed for all frames. For the remaining two sequences, a small number of camera poses were missing. This was due to failures in feature matching, which were primarily caused by motion blur from rapid camera movement. Importantly, the number of reconstructed 3D points was on the order of several tens of thousands for every sequence, and the overall shapes of gastric surfaces are sufficiently represented in the reconstructed point clouds, suggesting that subsequent NVS methods can be carried out in a meaningful manner. 
The mean reprojection errors, defined as the average errors in the pixel unit between image feature points and their reprojections from the reconstructed 3D points, range from 1.07221 to 1.36415. This indicates that the scenes were reconstructed with sufficient accuracy.
The set of the pinhole images, the camera poses, the camera's intrinsic parameters, and the point cloud are used as the inputs for subsequent NVS methods.

\begin{figure}[t]
\centering
\includegraphics[width=1\columnwidth]{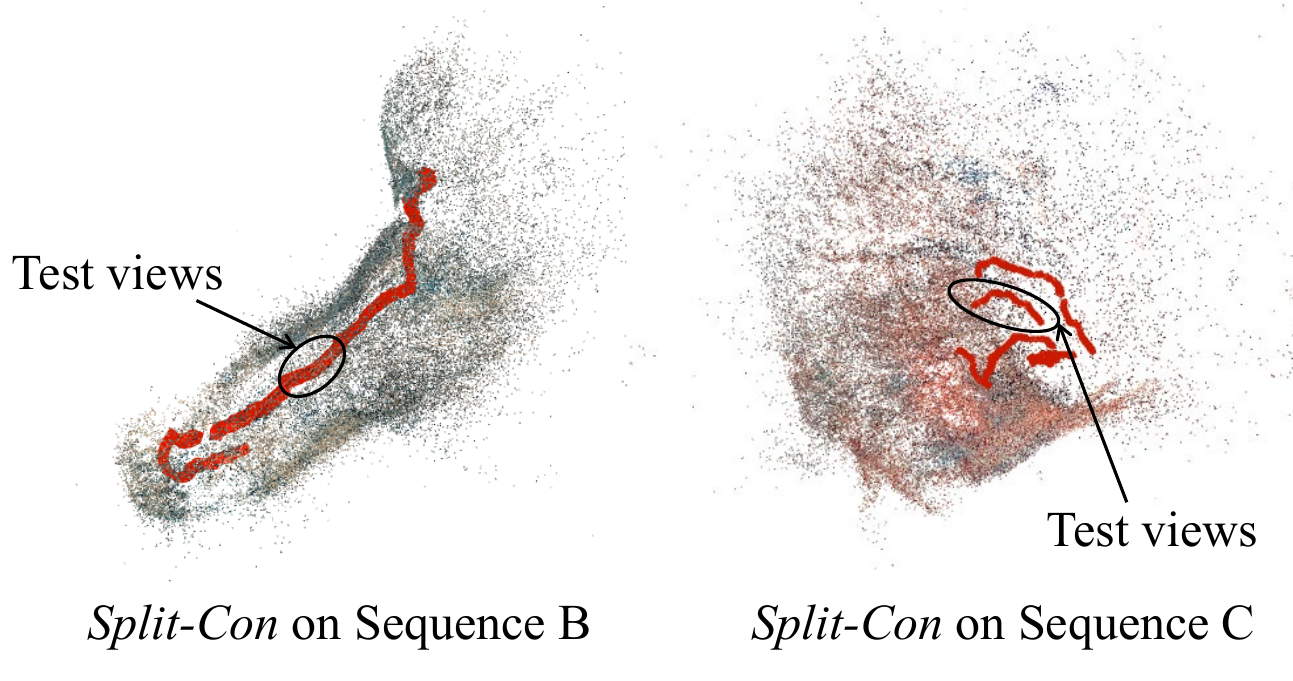}
\caption{Test views of \textit{Split-Con} on Sequence B and C.}
\label{fig3}
\end{figure}

For an effective evaluation of NVS methods, we prepared two types of train/test data splits for each sequence, both with a ratio of 7:1.
\begin{itemize}
    \item \textit{Split-Reg}: A data split where test views are sampled at regular intervals (one test view every eight frames), resulting in uniformly distributed test views throughout the sequence.
    \item \textit{Split-Con}: A data split where test views are sampled in a consecutive segment of the sequence, resulting in relatively large viewpoint differences between training and test views.
\end{itemize}
The first split, \textit{Split-Reg}, allows the model to be trained on images covering the entire sequence, and therefore, it is expected to yield high-quality NVS results. However, because the test views are spatially and temporally close to the training views, \textit{Split-Reg} makes it difficult to evaluate the model's performance for really ``novel" views. Thus, the second split, \textit{Split-Con}, as shown in Fig.~\ref{fig3}, is designed to evaluate the NVS performance for the viewpoints that are farther from the training views. Although this setting generally leads to lower NVS quality, \textit{Split-Con} enables a more meaningful evaluation for potential applications.

\section{Comparison of 3DGS-Based Methods}
To assess whether the dataset is suitable for NVS and to discuss the challenges of NVS for gastroendoscopy, we applied several 3DGS-based methods for comparisons, because the 3DGS-based methods achieve real-time rendering, while NeRF-based methods do not.

\subsection{Compared Methods}

Although the source code of some 3DGS-based methods for endoscopy is publicly available~\cite{zhu2024endogs,liu2024endogaussian,huang2024endo,zhao2024hfgs,bonilla2024gaussian,kaleta2025pr}, they are designed for the fixed viewpoint of the EndoNeRF dataset~\cite{zhu2024endogs,liu2024endogaussian,huang2024endo,zhao2024hfgs} or heavily rely on RNN-SLAM~\cite{ma2021rnnslam} for the initial reconstruction~\cite{bonilla2024gaussian,kaleta2025pr}. Thus, these methods are difficult to meaningfully evaluate and compare for our dataset. Instead, we selected general 3DGS-based methods frequently compared in the past computer vision literature, as summarized in Table~\ref{tab3}.

Specifically, we focused on smooth and textureless surfaces typical to stomach and selected the methods applying some geometric constraints, as well as the original 3DGS method~\cite{kerbl20233d}. The method termed 3DGS+depth extends the original 3DGS method~\cite{kerbl20233d} by incorporating an inverse depth map estimated by Depth Anything V2~\cite{yang2024depth} to compute a depth supervision loss during the optimization. By explicitly constraining the surface using the depth map, this approach improves geometric consistency. This method was implemented by the authors of 3DGS on their GitHub repository and released at the publication of Hierarchical 3DGS~\cite{kerbl2024hierarchical}. 2DGS~\cite{huang20242d} aims to improve surface representation quality by flattening Gaussians into disk representations. It further adopts a normal consistency term to ensure alignment between the geometries defined by depths and normals rendered from Gaussians. PGSR~\cite{chen2024pgsr} also flattens Gaussians and applies single-view and multi-view regularization using depth and normal renderings to enhance geometric consistency. Finally, GSDF~\cite{yu2024gsdf} explicitly models surfaces by jointly learning a signed distance field (SDF) representation \cite{wang2021neus} together with the 3D Gaussian representation while enforcing depth and normal consistency between them.

\begin{table}[t]
\centering
\caption{Properties of the compared 3DGS-based methods.}
\label{tab3}
\resizebox{\columnwidth}{!}{
\begin{tabular}{lccccc}
\hline
Methods &
\makecell{Depth\\supervision} &
\makecell{Depth/Normal\\consistency} &
\makecell{Flattening\\Gaussians} &
\makecell{SDF\\representation} \\
\hline
3DGS    &  &  &  & \\
3DGS+depth & \checkmark &  &  & \\
2DGS    &  & \checkmark & \checkmark &  \\
PGSR    &  & \checkmark & \checkmark &  \\
GSDF    &  & \checkmark &  & \checkmark \\
\hline
\end{tabular}}
\end{table}

All methods were evaluated using their default settings. The methods other than GSDF were trained for 30,000 iterations, while GSDF was trained for 45,000 iterations because it first optimizes the Gaussian representation for 15,000 iterations to establish the coarse geometry and then incorporates the SDF representation for the remaining 30,000 iterations. The 3D Gaussians in all methods were initialized with the same filtered point cloud obtained after PGSR's preprocessing step~\cite{chen2024pgsr} that removes SfM points with large reprojection errors or low observation counts.

\subsection{Results}

\begin{table*}[t!]
\centering
\small
\caption{Quantitative comparison on \emph{Split-Reg}. Higher is better for PSNR and SSIM, while lower is better for LPIPS.}
\label{tab4}
\renewcommand{\arraystretch}{1.15}
\setlength{\tabcolsep}{4pt}

\begin{tabular}{ccc}

\begin{tabular}{lccc}
\hline
\multicolumn{4}{c}{\textbf{A (65 test views)}} \\
\hline
Method & PSNR & SSIM & LPIPS \\
\hline
3DGS         & 27.08 & 0.734 & 0.463 \\
3DGS+depth   & 27.06 & 0.733 & 0.461 \\
2DGS         & 25.52 & 0.714 & 0.500 \\
PGSR         & 25.72 & 0.717 & 0.480 \\
GSDF         & \textbf{27.17} & \textbf{0.736} & \textbf{0.443} \\
\hline
\end{tabular}
&
\begin{tabular}{lccc}
\hline
\multicolumn{4}{c}{\textbf{B (55 test views)}} \\
\hline
Method & PSNR & SSIM & LPIPS \\
\hline
3DGS         & \textbf{26.87} & 0.834 & 0.424 \\
3DGS+depth   & 26.78 & \textbf{0.836} & \textbf{0.418} \\
2DGS         & 24.96 & 0.816 & 0.463 \\
PGSR         & 24.82 & 0.814 & 0.447 \\
GSDF         & 25.76 & 0.824 & 0.422 \\
\hline
\end{tabular}
&
\begin{tabular}{lccc}
\hline
\multicolumn{4}{c}{\textbf{C (46 test images)}} \\
\hline
Method & PSNR & SSIM & LPIPS \\
\hline
3DGS         & 21.87 & 0.654 & 0.402 \\
3DGS+depth   & 21.97 & \textbf{0.659} & 0.397 \\
2DGS         & 21.28 & 0.636 & 0.444 \\
PGSR         & 21.55 & 0.640 & 0.420 \\
GSDF         & \textbf{22.22} & 0.650 & \textbf{0.391} \\
\hline
\end{tabular}

\\[1.5ex]

\begin{tabular}{lccc}
\hline
\multicolumn{4}{c}{\textbf{D (33 test images)}} \\
\hline
Method & PSNR & SSIM & LPIPS \\
\hline
3DGS         & \textbf{27.55} & 0.854 & 0.331 \\
3DGS+depth   & 27.24 & \textbf{0.854} & \textbf{0.326} \\
2DGS         & 25.49 & 0.834 & 0.372 \\
PGSR         & 26.16 & 0.841 & 0.354 \\
GSDF         & 26.83 & 0.847 & 0.334 \\
\hline
\end{tabular}
&
\begin{tabular}{lccc}
\hline
\multicolumn{4}{c}{\textbf{E (54 test images)}} \\
\hline
Method & PSNR & SSIM & LPIPS \\
\hline
3DGS         & 25.42 & 0.717 & 0.464 \\
3DGS+depth   & 25.60 & \textbf{0.719} & 0.458 \\
2DGS         & 24.05 & 0.697 & 0.499 \\
PGSR         & 24.28 & 0.695 & 0.483 \\
GSDF         & \textbf{25.84} & 0.718 & \textbf{0.441} \\
\hline
\end{tabular}
&
\begin{tabular}{lccc}
\hline
\multicolumn{4}{c}{\textbf{Average of A--E}} \\
\hline
Method & PSNR & SSIM & LPIPS \\
\hline
3DGS         & 25.74 & 0.759 & 0.417 \\
3DGS+depth   & \textbf{25.73} & \textbf{0.760} & 0.412 \\
2DGS         & 24.26 & 0.739 & 0.456 \\
PGSR         & 24.51 & 0.741 & 0.437 \\
GSDF         & 25.56 & 0.755 & \textbf{0.406} \\
\hline
\end{tabular}

\end{tabular}
\end{table*}

\begin{figure*}[t!]
\centering
\includegraphics[width=0.94\textwidth]{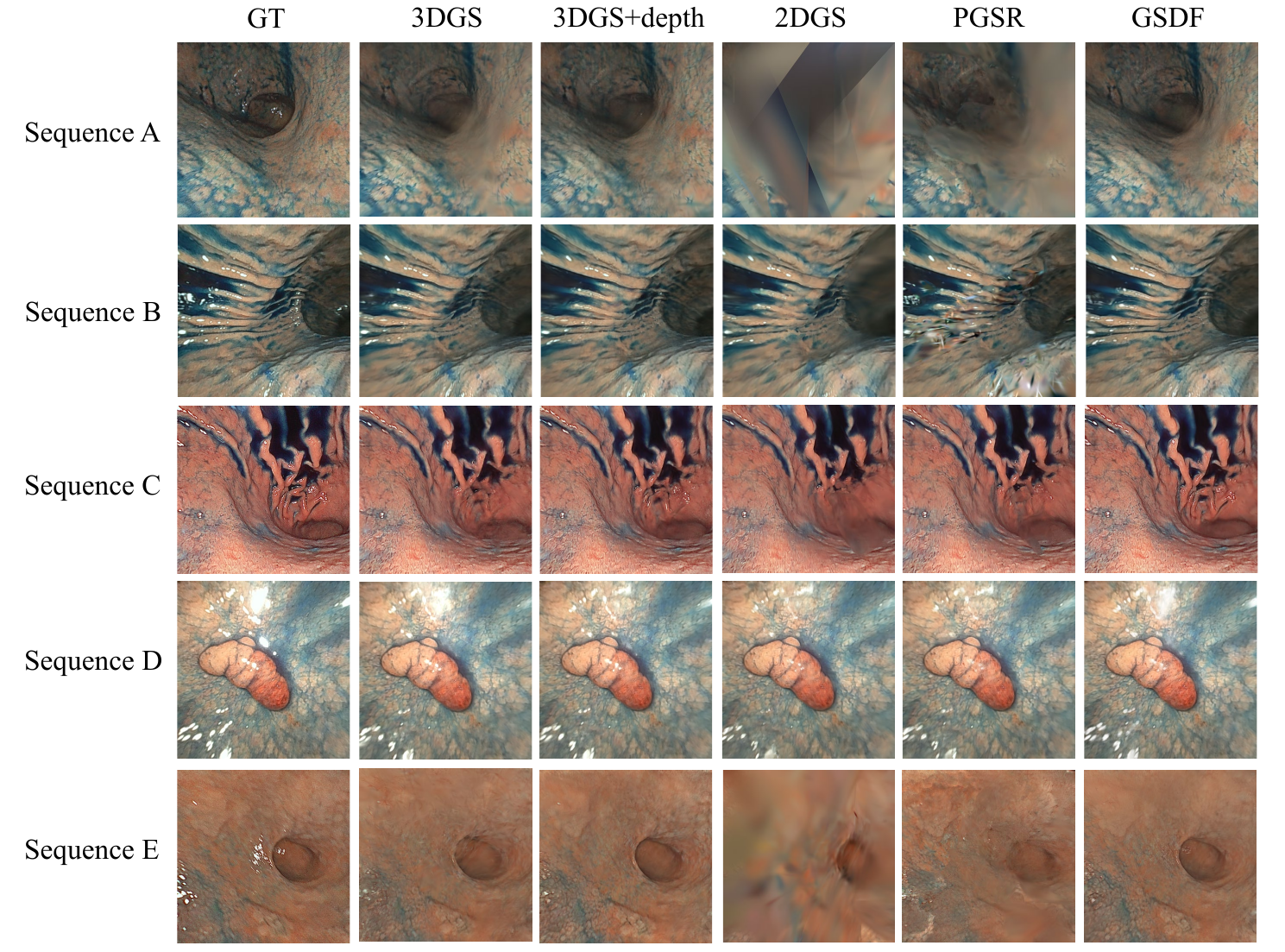}
\caption{Visual comparisons for the test views of \emph{Split-Reg}.}
\label{fig4}
\end{figure*}

\begin{table*}[t!]
\centering
\small
\caption{Quantitative comparison on \emph{Split-Con}. Higher is better for PSNR and SSIM, while lower is better for LPIPS.}
\label{tab5}
\renewcommand{\arraystretch}{1.15}
\setlength{\tabcolsep}{4pt}

\begin{tabular}{ccc}

\begin{tabular}{lccc}
\hline
\multicolumn{4}{c}{\textbf{A (65 test views)}} \\
\hline
Method & PSNR & SSIM & LPIPS \\
\hline
3DGS         & 18.16 & 0.610 & 0.572 \\
3DGS+depth   & 18.66 & 0.612 & 0.567 \\
2DGS         & 16.84 & 0.597 & 0.603 \\
PGSR         & 17.39 & 0.598 & 0.587 \\
GSDF         & \textbf{19.84} & \textbf{0.619} & \textbf{0.544} \\
\hline
\end{tabular}
&
\begin{tabular}{lccc}
\hline
\multicolumn{4}{c}{\textbf{B (55 test views)}} \\
\hline
Method & PSNR & SSIM & LPIPS \\
\hline
3DGS         & 17.38 & 0.662 & 0.516 \\
3DGS+depth   & 19.62 & 0.659 & 0.484 \\
2DGS         & 16.90 & 0.644 & 0.545 \\
PGSR         & 17.39 & 0.639 & 0.527 \\
GSDF         & \textbf{19.71} & \textbf{0.670} & \textbf{0.465} \\
\hline
\end{tabular}
&
\begin{tabular}{lccc}
\hline
\multicolumn{4}{c}{\textbf{C (46 test images)}} \\
\hline
Method & PSNR & SSIM & LPIPS \\
\hline
3DGS         & 16.69 & 0.450 & 0.533 \\
3DGS+depth   & 16.71 & 0.449 & 0.531 \\
2DGS         & 16.67 & 0.467 & 0.548 \\
PGSR         & 16.53 & 0.450 & 0.547 \\
GSDF         & \textbf{17.62} & \textbf{0.476} & \textbf{0.504} \\
\hline
\end{tabular}

\\[1.5ex]

\begin{tabular}{lccc}
\hline
\multicolumn{4}{c}{\textbf{D (33 test images)}} \\
\hline
Method & PSNR & SSIM & LPIPS \\
\hline
3DGS         & 18.40 & 0.694 & 0.466 \\
3DGS+depth   & 20.10 & 0.714 & \textbf{0.419} \\
2DGS         & 17.29 & 0.688 & 0.524 \\
PGSR         & 17.28 & 0.684 & 0.520 \\
GSDF         & \textbf{20.48} & \textbf{0.718} & 0.421 \\
\hline
\end{tabular}
&
\begin{tabular}{lccc}
\hline
\multicolumn{4}{c}{\textbf{E (54 test images)}} \\
\hline
Method & PSNR & SSIM & LPIPS \\
\hline
3DGS         & 19.32 & 0.640 & 0.541 \\
3DGS+depth   & \textbf{21.03} & \textbf{0.652} & 0.521 \\
2DGS         & 18.40 & 0.624 & 0.570 \\
PGSR         & 19.02 & 0.622 & 0.565 \\
GSDF         & 20.92 & 0.642 & \textbf{0.498} \\
\hline
\end{tabular}
&
\begin{tabular}{lccc}
\hline
\multicolumn{4}{c}{\textbf{Average of A--E}} \\
\hline
Method & PSNR & SSIM & LPIPS \\
\hline
3DGS         & 17.99 & 0.611 & 0.526 \\
3DGS+depth   & 19.22 & 0.617 & 0.504 \\
2DGS         & 17.22 & 0.604 & 0.558 \\
PGSR         & 17.52 & 0.599 & 0.549 \\
GSDF         & \textbf{19.71} & \textbf{0.625} & \textbf{0.486} \\
\hline
\end{tabular}

\end{tabular}
\end{table*}

\begin{figure*}[t!]
\centering
\includegraphics[width=0.95\textwidth]{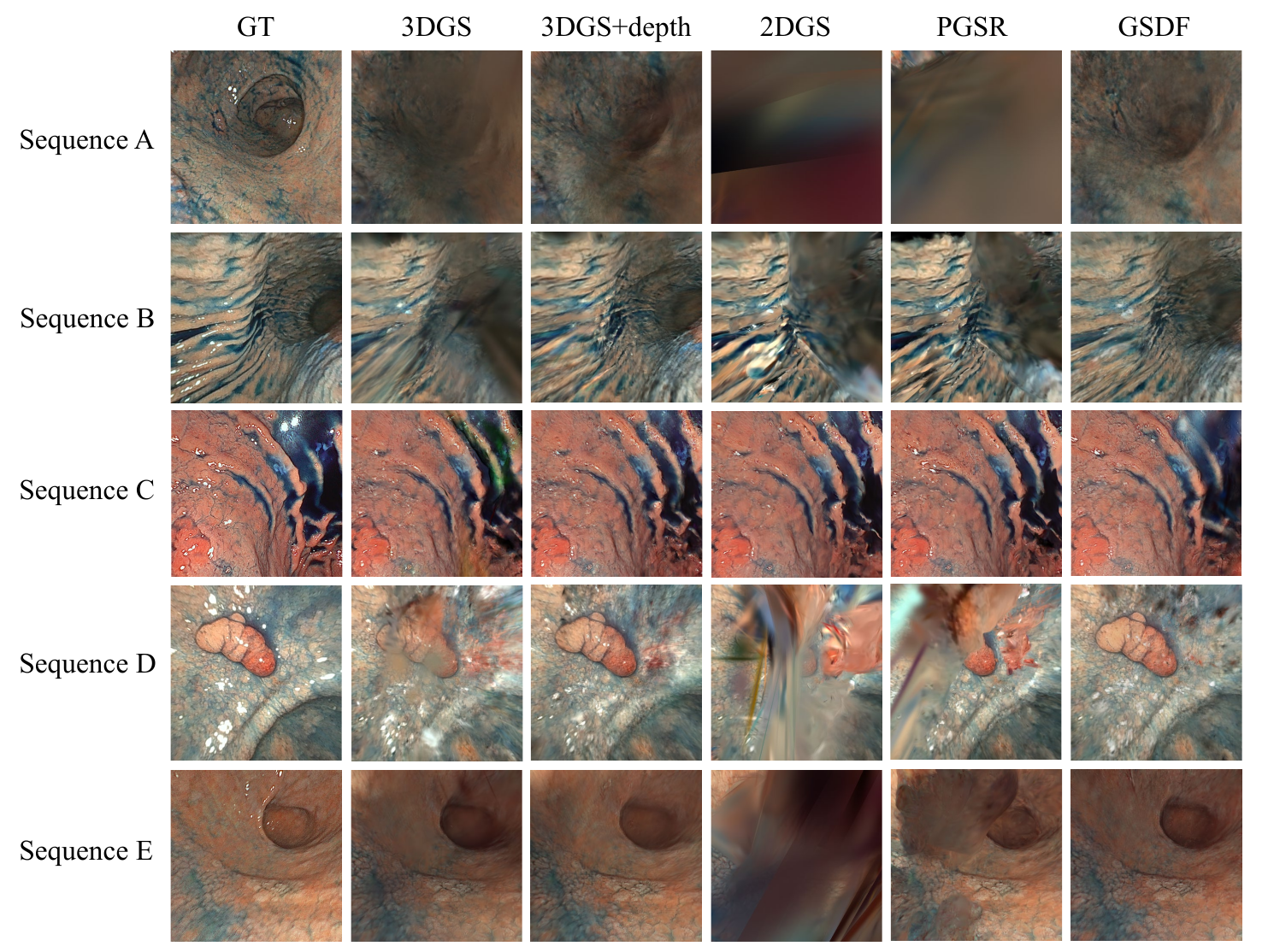}
\caption{Visual comparison for the test view of \emph{Split-Con}.}
\label{fig5}
\end{figure*}

\subsubsection{Split-Reg}
Table~\ref{tab4} shows the quantitative comparison. We can see that the methods that explicitly optimize geometry, i.e., 3DGS+depth and GSDF, consistently achieve better metrics, with minor variations across sequences. In contrast, 2DGS and PGSR, which flatten Gaussians, produce consistently lower scores than the other methods across all metrics. The visual comparisons, shown in Fig.~\ref{fig4}, are consistent with the quantitative results reported in the table. In particular, although 2DGS produces reasonable reconstructions in some scenes, it frequently generates flattened Gaussians at inconsistent and erroneous locations, which locally obstruct the view. Similarly, PGSR often exhibits noisy reconstructions, as observed in Sequences A, B, and E, where surfaces are generated at locations that deviate from the true underlying geometry. These limitations can be attributed to the fact that, unlike the other methods, these two methods adopt flat Gaussian representations, which are insufficient to model the curved structures of the gastric surfaces as well as subtle surface deformations.
In contrast, 3DGS yields reconstructions in which the overall surface structure is recognizable, despite minor artifacts such as blurred regions in the background of Sequence B and the presence of dark floating artifacts in Sequence E. This indicates that 3DGS exhibits limited capability of learning accurate geometry.
On the other hand, both 3DGS+depth and GSDF effectively suppress the floating artifacts and blur observed in 3DGS, as 3DGS+depth leverages depth maps and GSDF employs an SDF-based surface representation to encourage the optimization of Gaussians to align with the underlying geometry. Also, these two methods achieve consistently high-quality renderings that preserve fine details.

\subsubsection{Split-Con}
Table~\ref{tab5} shows the quantitative comparison. Since the rendered results are generated from viewpoints that are distant from the training views, the overall evaluation scores are substantially lower than those of \emph{Split-Reg}, as expected. Unlike the results observed in \emph{Split-Reg}, GSDF achieves the best performance across all evaluation metrics. Similarly to the results of \emph{Split-Reg}, 2DGS and PGSR consistently exhibit inferior performance. From the visual comparisons shown in Fig.~\ref{fig5}, the performance differences among the compared methods become more obvious than in that of \emph{Split-Reg}. 3DGS, 2DGS, and PGSR exhibit prominent artifacts, and in many views, even the coarse structure cannot be reliably perceived. In contrast, although minor blur and floating artifacts remain, 3DGS+depth and GSDF achieve robust renderings even from viewpoints far from the training views.

From the results on Sequence E, a phenomenon specific to endoscopic sequences can be observed, namely that the rendered outputs exhibit significant brightness differences compared to the ground truth. As can be observed from the original images shown in Fig.~\ref{fig2}, this issue arises from inter-view illumination inconsistencies caused by variations in the lighting direction and intensity. In turn, they lead to degraded reconstruction quality. To address this challenge, it would be desirable to develop methods that explicitly account for illumination variations, like the very recent CollaGS\cite{wang2025moving}.

\subsubsection{Video results}
In our project page, we provide a supplementary video showing the rendering results (trained using the train views of \emph{Split-Reg}) along a shared camera trajectory that is different from both the training and the test views. The video results demonstrate the current performance of each method for user-interactive NVS.

\section{Conclusion}
In this paper, we presented GastroNVS, the first gastroendoscopic dataset specifically designed for NVS inside the stomach. The dataset consists of real gastroendoscopic image sequences which covers wide regions of the stomach under realistic clinical conditions, accompanied by SfM results that sufficiently represent the structures.
From the experimental results of some representative 3DGS-based methods on two types of train/test split, we observed that the methods explicitly learning surface geometry are more effective for NVS in the gastroendoscopic scene. Moreover, the results revealed an endoscopy-specific challenge caused by inter-view illumination inconsistency, which leads to significant brightness discrepancies and degraded reconstruction quality in a certain sequence.
We believe that GastroNVS is a valuable dataset for gastroendoscopic NVS, with the potential to facilitate the development of techniques that may support clinical navigation and diagnosis in the future.

\bibliographystyle{IEEEtran}
\bibliography{egbib}

\end{document}